# Attention to Detail: Fine-Scale Feature Preservation-Oriented Geometric Pre-training for AI-Driven Surrogate Modeling

Yu-hsuan Chen[1]    Jing Bi[2]    Cyril Ngo Ngoc[2]

Victor Oancea[2]    Jonathan Cagan[1]    Levent Burak Kara[1]

[1]Carnegie Mellon University, 5000 Forbes Avenue, Pittsburgh, PA, USA*
[2]Dassault Systèmes SIMULIA Corp., 1301 Atwood Avenue, Johnston RI, USA

**ABSTRACT**

*AI-driven surrogate modeling has become an increasingly effective alternative to physics-based simulations for 3D design, analysis, and manufacturing. These models leverage data-driven methods to predict physical quantities traditionally requiring computationally expensive simulations. However, the scarcity of labeled CAD-to-simulation datasets has driven recent advancements in self-supervised and foundation models, where geometric representation learning is performed offline and later fine-tuned for specific downstream tasks. While these approaches have shown promise, their effectiveness is limited in applications requiring fine-scale geometric detail preservation. This work introduces a self-supervised geometric representation learning method designed to capture fine-scale geometric features from non-parametric 3D models. Unlike traditional end-to-end surrogate models, this approach decouples geometric feature extraction from downstream physics tasks, learning a latent space embedding guided by geometric reconstruction losses. Key elements include the essential use of near-zero level sampling and the innovative batch-adaptive attention-weighted loss function, which enhance the encoding of intricate design features. The proposed method is validated through case studies in structural mechanics, demonstrating strong performance in capturing design features and enabling accurate few-shot physics predictions. Comparisons with traditional parametric surrogate modeling highlight its potential to bridge the gap between geometric and physics-based representations, providing an effective solution for surrogate modeling in data-scarce scenarios.*

## 1. INTRODUCTION

Recent advancements in AI-driven surrogate models are becoming increasingly more effective at providing an alternative to physics-based simulations for 3D design, analysis, and manufacturing [1–5]. These models leverage data-driven methods to efficiently predict physical quantities of interest, traditionally requiring computationally expensive simulations. However, the challenge of acquiring universally useful and widely applicable labeled CAD-to-simulation data pairs —has led to the recent emergence of self-supervision and foundation models. In these approaches, offline learning focuses primarily on accurate geometric representation learning, which is later fine-tuned for specific downstream tasks using a limited set of labeled data [5–7]. While these methods have demonstrated promise, particularly when pretrained on large-scale unlabeled CAD repositories such as FabWave [8], MFCAD [9], and Fusion 360 [10], their primary applications have been in downstream engineering tasks that exhibit only moderate sensitivity to fine-scale geometric features, such as classification, segmentation, and single value regression.

However, in mechanical product design, many simulations such as structural stress analysis, buckling calculations, and deformation modeling, rely on physics that are highly sensitive to small-scale geometric details [11–13]. Features such as tight fillets, thin-shelled structures, and intricate stylistic or functional surface details can significantly impact mechanical performance. Existing geometric representation learning techniques often fail to preserve these crucial fine-scale features, leading to a loss of fidelity in surrogate models. This limitation is particularly problematic when attempting to learn mappings from geometric representations to simulation results, as inadequate geometric detail retention can result in erroneous predictions and unreliable surrogate modeling [14].

While parametric CAD models could alleviate this by providing a high-fidelity geometric representation of small-scale features, generating parametric models for large and diverse corpora of existing CAD repositories is largely infeasible [15]. Most CAD repositories contain non-parametric models such as mesh and boundary representations, making it highly desirable to develop pre-trained geometric representation models that can operate on non-parametric geometries. Despite the clear benefits, no existing methods adequately address this challenge, creating a critical gap in the field of geometric surrogate modeling.

To bridge this gap, a geometric representation learning method is introduced in this work that prioritizes fine-scale detail preservation through a self-supervised pretraining phase. Unlike traditional end-to-end surrogate modeling approaches, this method separates geometric feature extraction from downstream simulation prediction tasks. Using a training corpus of non-parametric, boundary representation (B-Rep) models as input, a fixed-length latent space embedding of 3D geometries is learned. The latent vector effectively functions as a learned parametric representation of the input non-parametric geometries. This

---

representation is guided exclusively by geometric reconstruction losses, eliminating the need for a priori simulation data. The pre-trained model then serves as a foundation for downstream surrogate modeling on two time-evolving physics tasks of reaction force and deformation field prediction during an axial load process. The key elements that enable precise encoding of intricate design features in geometric representation learning are the essential use of near-zero level sampling, periodic spatial locality indicator, and the innovative batch-adaptive attention-weighted loss function. By leveraging this pretraining strategy, the reliance on large-scale labeled simulation datasets is significantly reduced, allowing direct usage of pretrained latent representation and fine-tuning with minimal supervision while maintaining high accuracy in predicting time-dependent force vector and scalar fields.

The effectiveness of this method is validated through two case studies. The first focuses on a crash box featuring thin-walled structures, where thickness variations play a crucial role in impact resistance. The second examines a bottle design, where detailed rib geometries influence mechanical performance. The pre-trained latent vectors successfully capture fine-scale design features, as demonstrated by strong linear probing performance on design parameters and accurate few-shot predictions of reaction forces and nodal displacement fields. Additionally, comparisons against traditional parametric surrogate modeling highlight the extent to which pretraining on non-parametric data can bridge the gap between purely geometric representations and explicit parametric definitions.

To establish a rigorous performance benchmark, both case studies employ parametric designs where standalone parameter-to-simulation networks are trained, serving as an upper-bound performance reference. However, this comparison is conducted solely for evaluation purposes, as parametric models are not assumed to be available when deploying AI-driven surrogate models in real-world scenarios.

The main contributions of this work are as follows:

1. A self-supervised representation learning framework that automatically discovers and extracts varying design parameters from non-parametric geometries without requiring predefined parameter information.

2. A set of key techniques necessary for effectively capturing fine-scale geometric parameters.

3. A learned latent space that enables accurate structural physics estimation, outperforming fully supervised learning in few-shot scenarios, along with insights into how different downstream tasks influence the performance of direct latent vector deployment versus encoder fine-tuning.

## 2. RELATED WORK

This research seeks to enhance geometric representation learning (GRL) methods to better capture fine-scale yet critical design features essential for structural physics applications, aiming to achieve performance levels comparable to parametric surrogate modeling when available. This section reviews various data-driven surrogate models, including parametric, non-parametric, and self-supervised approaches. It then narrows the focus to self-supervised learning in CAD modalities, concluding with an examination of prior work in implicit shape reconstruction.

**Parametric surrogate models,** which use design parameters as inputs to predict desired quantities and demonstrate success across various engineering design applications [16]. Neural networks are commonly used as approximators due to their flexibility in handling complex input-output relationships, leveraging parallel computing, and enhancing computational efficiency. For instance, Mai et al. [1] used the cross-sectional areas of truss members as inputs to predict nodal displacements under nonlinear deformation. Mozaffar et al. [17] developed a recurrent neural network model that takes non-temporal microstructure descriptors and temporal deformation paths as inputs to predict temporal stresses and plastic energy over 100 increments for each representative volume elements, enabling precise and efficient material plasticity predictions. Other studies leverage frequency and mode shape inputs to detect damage locations and severity in planar trusses [18,19]. Similarly, Ribeiro et al. [20] predicted internal pressure in reinforced panels based on width, length, and other design parameters, achieving results 40 times faster than commercial finite element methods while maintaining acceptable accuracy. Additionally, Kazeruni and Ince [21] used material properties and elastic stress inputs to estimate actual elastic-plastic stress and strain fields for fatigue failure analysis of notched bodies.

Despite these advancements, many engineering applications lack the necessary parametric data due to challenges in data formulation or the absence of labeled geometries. Furthermore, the variable length of input data often limits the generalizability of these models. Comprehensive reviews of parametric surrogate models discuss additional engineering applications and challenges in detail [16,22].

**Non-parametric approaches** have emerged as powerful alternatives for both 2D and 3D problems, demonstrating that, given sufficiently large datasets, convolutional neural networks (CNNs) and graph neural networks (GNNs) can accurately predict complex scalar stress fields [2,23–25]. Nie et al. [23] and Jiang et al. [23] utilized pixelated image data to predict stress distributions in 2D structures, while Ferguson et al. [25] employed interpolated multi-resolution CNNs capable of operating on arbitrary meshes, achieving higher-fidelity results. Another line of research leverages voxel representations combined with 3D CNNs to predict manufacturing costs [26,27], production time [28], and other design-relevant quantities [29].

Despite their effectiveness, these methods are constrained by their reliance on simulation results formatted as rasterized modalities, limiting resolution flexibility and introducing additional preprocessing overhead [15]. TAG U-Net, a graph convolutional network, was developed to handle diverse mesh and graph structures, predicting displacement fields induced by thermal residual stress directly from nodal data without requiring image-based input. Trained and validated on a 3D additive manufacturing dataset, it achieved a median R-squared value exceeding 0.85 on test cases. Other approaches focus on point-to-point mappings, such as Transolvers [3], which employed multi-attention mechanisms for various physics applications, and MeshGraphNet [4], designed for dynamically evolving graphs. However, these models rely on large volumes of accurately labeled simulation data, which can be costly and time-intensive to obtain.

**Representation learning**, particularly through self-supervised learning, provides a powerful framework that eliminates the need for labeled data, making it applicable across diverse tasks and modalities. In image processing, self-supervised representation learning has enabled CNNs and transformer models to achieve faster and superior classification accuracy through various techniques, such as inpainting [30,31], rotation prediction [32], and contrastive learning [33]. These methods have demonstrated that self-supervised learning not only matches or outperforms training non-parametric models from scratch on downstream tasks but also facilitates few-shot learning, allowing models to generalize patterns with minimal training samples. A similar trend is observed in natural language processing, where masked language models, trained to predict missing tokens based on contextual cues, have achieved state-of-the-art results on benchmarks such as SQuAD, GLUE, and MNLI [34,35]. Another widely used approach involves pre-training models for next-word prediction, which has led to significant advancements in generative performance, exemplified by models such as ChatGPT [34,36,37]. However, applying these techniques to design workflows—particularly when dealing with vectorized Computer-Aided Design (CAD) data or non-Euclidean datasets—raises a critical question: do these methods retain their effectiveness in these contexts?

**Self-supervised representation learning for CAD structures** has made significant strides, particularly in the application of GNNs to both 2D and 3D domains. In 2D vector graphics (SVGs), GNNs are applied to tasks such as manufacturing classification, engineering drawing segmentation, and stylized feature extraction [38–40]. In the 3D domain, researchers have primarily modeled Boundary Representation (B-Rep) structures as hierarchical graphs, where surfaces form nodes interconnected by curves [5–7,10,14]. One key advancement is UV-Net [7], which parameterizes geometries using the U and V domains of curves and surfaces. This method demonstrates superior performance in part classification and surface segmentation, leveraging a Siamese contrastive learning framework for self-supervised learning, with promising results in shape retrieval. Building on this, Jones et al. [14] introduced an approach that maps predefined graph features to reconstruct the surface signed distance field (SDF) in the UV domain, translating explicit surface features into implicit boundary representations. This improvement significantly boosted few-shot learning performance, establishing new benchmarks in classification and segmentation tasks. More recently, VIRL represents a novel approach in self-supervised representation learning for Computer-Aided Manufacturing (CAM) and Finite Element Analysis (FEA) simulation surrogates [5]. VIRL incorporates spatial information during pretraining by reconstructing geometries' volumetric SDF, enhancing the GNN encoder's understanding of shape representations and spatial awareness. Despite promising results, the $R^2$ values remain at a proof-of-concept stage, with the method focusing on single-value regression, limiting its ability to capture more complex field distributions relevant to practical engineering applications.

These methods highlight the potential of self-supervised learning in CAD, but the comparison to parametric surrogate models remains underexplored, particularly due to the absence of design parameters in existing CAD repositories. Furthermore, when dealing with geometries that demand high precision or involve challenging structures such as thin shells, previous reconstruction methods often fall short. Given that the accuracy of downstream learning tasks correlates with the accuracy of a model's reconstruction [14], failing to capture these critical details significantly limits the applicability of previous methods, especially in structural physics applications where fine-scale yet essential shape features strongly influence simulation results.

**Implicit shape reconstruction** is a key challenge when mapping from an implicit latent space to an SDF, as it requires achieving high-fidelity reconstructions without the constraints of voxel discretization. DeepSDF [41] lays a strong foundation by learning a continuous implicit function for shape representation, enabling smooth and high-quality reconstructions. Advancements such as SIREN [42], FFN [43], and BACON [44] further improve upon this by introducing periodic transformations to capture fine details and high-frequency variations in geometry. However, these approaches are generally designed for single-data reconstruction, leaving uncertainty about their ability to generalize to dynamically changing inputs or produce meaningful and structured latent representations for downstream tasks.

Consequently, while existing self-supervised methods for CAD have made progress in few-shot learning capabilities—removing the restriction of needing predefined design parameters and surpassing non-parametric methods in performance—they struggle to capture fine-scale details crucial for engineering tasks such as stress field estimation and deformation modeling. Therefore, an improved approach that prioritizes the accurate representation of intricate geometric features is needed to better support applications requiring high precision, such as structural analysis and material behavior prediction.

## 3. METHODS

Fully supervised learning methods require large amounts of data, as the model must simultaneously learn a latent representation and decode it for downstream tasks. To mitigate this challenge, a two-stage training strategy is adopted. Rather than relying directly on extensive simulation data for task-specific learning, the first stage leverages multiple representations of CAD data, which is readily available and does not require solving computationally expensive PDEs. Using an encoder-decoder structure, this pre-training process maps one representation (B-Rep) to another (SDF), enabling the encoder to learn a latent space that closely aligns with the underlying design parameters. This approach is expected to reduce the physics simulation data required for downstream task performance compared to training non-parametric surrogate models from scratch.

Recent efforts have explored self-supervised learning on CAD geometries to extract informative latent representations for few-shot applications, aiming to circumvent the data-intensive nature of fully supervised surrogate models. However, a fundamental challenge remains unaddressed: existing methods struggle to capture design parameters that control fine-scale geometric features. This limitation significantly impacts domains where precise geometry plays a critical role, such as structural physics estimation. Addressing this challenge requires a pretraining strategy that explicitly guides the encoder towards capturing geometric variations that drive meaningful differences in downstream tasks.

The proposed methodology consists of three key components: (1) case study selection and geometry creation, (2) B-Rep to SDF pretraining, and (3) downstream task learning. The core insight behind the pretraining approach is that accurately reconstructing the geometries' signed distance fields (SDFs) enables the encoder to learn a structured and meaningful latent representation.

This section begins by introducing the selection of case studies to address the specific challenges posed by these geometries. It then outlines the pre-training architecture and demonstrates how fine-scale design parameters are effectively captured, with comparisons to previous approaches. The section concludes with a description of the downstream task simulation setup and the methodology used for evaluating the proposed method.

### 3.1 Case Study Selection and Data Creation

The geometries in this study are generated using parametric modeling systems and sourced from Dassault Systèmes [45]. Two representative datasets, crash boxes and bottles, are constructed to evaluate different aspects of the pretraining framework.

### 3.1.1 Crash Box Dataset

The crash box is a structural component installed at the front of a vehicle, playing a critical role in absorbing impact energy during collisions [46]. The crash box geometries in the dataset are parameterized by four design variables: height, width, length, and thickness, as illustrated in Figure 1. Although actual crash box designs are considerably more complex, the simplified geometry offers a reasonable approximation for evaluating deformation under external forces, a critical factor for automotive safety.

The second dataset consists of bottle geometries, with a focus on fine structural details that influence deformation under top load. The dataset is controlled by four key parameters affecting the upper half of the bottle: rib thickness, rib pitch, rib spacing, and top section radius (see Figure 2). These parameters determine both local geometric features and global structural behavior. Studying how bottles deform under compression is critical for ensuring stability during transportation and storage, where stacked bottles experience forces that can lead to structural failure [47]. The distribution and geometry of the rib structures play a key role in reinforcing the bottle's resistance to these compressive loads by redistributing stress and preventing localized buckling. Optimizing bottle design can lead to stronger yet lightweight structures, reducing material usage while maintaining durability. Additionally, improved stacking stability minimizes packaging failures, enhances logistical efficiency, and reduces waste—making this study highly relevant for both economic and sustainability considerations in the packaging industry.

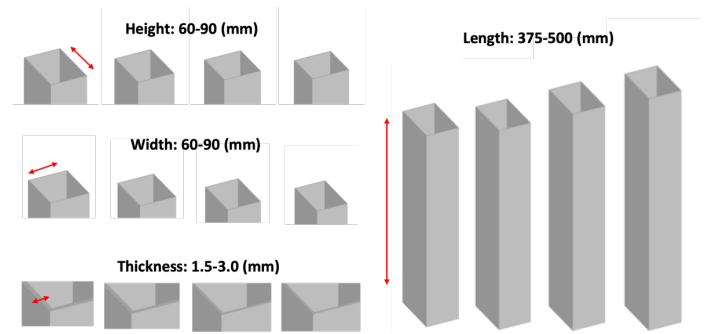

**Figure 1:** The crash box geometry variations are caused by height, width, length, and thickness.

### 3.1.2 Bottle Dataset

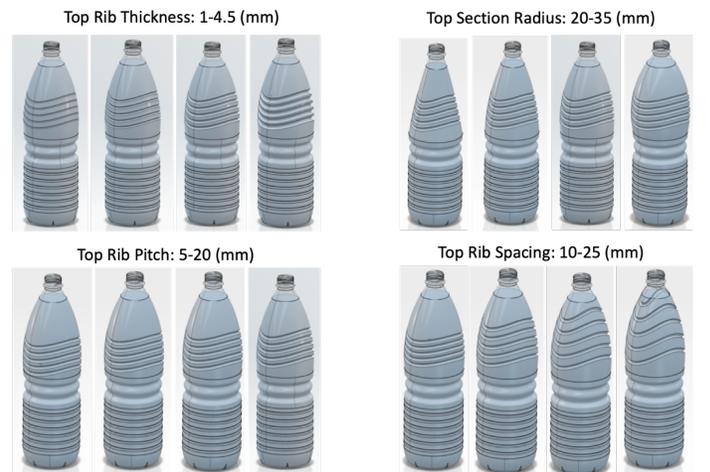

**Figure 2.** The bottle geometry variations caused by thickness, radius, pitch, and spacing of ribs.

For both datasets, design space is systematically sampled using 4 design parameters per shape, with each parameter discretized into six evenly spaced values, resulting in 1,296 unique geometries. Of these, 1,200 data are randomly allocated for training and reserve the remaining for testing.

### 3.1.3 Motivation for Case Study Selection

The goal of this study is to explore whether and how self-supervised geometric representation learning can capture small-scale geometry features governed by design parameters, a challenge often overlooked in prior domain-specific works. Therefore, the selection of two specific geometries is driven by their practical significance and the technical challenges associated with Signed Distance Field (SDF)-based representation learning:

- **Application Relevance** – Both geometries play key roles in structural deformation. Crash boxes, designed to absorb extreme impact forces in automotive applications, and bottles, which must maintain integrity under compressive loads during stacking and transport, provide critical constraints for evaluating learned geometric representations.
- **Thin Shell Structures (crash box)** – Traditional SDF representations face challenges with thin-walled geometries like crash boxes, where the sparsity of negative signed distance values (inside the shape) complicates accurate shape capture. Crash boxes primarily consist of a thin outer shell with minimal solid interior, making uniform-grid-based SDF interpolation ineffective. This study examines whether SDF-informed pre-training, with specific techniques, can effectively encode the variation of the small scale thickness design parameter.
- **Complex, High-Detail Surfaces (bottles)** – While previous CAD representation learning methods have succeeded with simpler geometries, they often fail to capture intricate, high-frequency details. Bottles, with their fine rib structures and varying top profiles, present a challenging test case to assess whether SDF-informed pre-training can recreate these fine geometric details through structured latent representations.

### 3.2 Data Preparation and Pre-training Architecture

Boundary representation (B-Rep) is a natural and widely used input modality in human-centered design systems. Since CAD models are typically created using B-rep, this representation is particularly relevant when structural physics insights are needed for designers to optimize and refine geometries. Consequently, developing a model that directly takes B-rep as input would be highly beneficial. However, B-rep data presents significant challenges for neural network processing due to its inherent inhomogeneity:

- Models contain varying numbers of surfaces.
- Surfaces contain varying numbers of types, and are connected to varying numbers of curves.
- Curves contain varying numbers of types, and are connected to varying numbers of vertices.

These characteristics make B-rep incompatible with models requiring fixed-topology inputs. To address this, B-rep is commonly preprocessed into hierarchical graphs and processed using hierarchical graph neural networks (GNNs). This study employs a parameterized encoding strategy, where curves and surfaces are sampled at fixed vertex counts, with coordinates and normal vectors of sampled points serving as features. This approach captures geometric details while maintaining expressiveness across surface types, making it suitable for complex designs like the bottle dataset. Figure 3 illustrates the encoded crash box and bottle datasets, highlighting the complexity contrast: crash boxes contain dozens of edges and faces, while bottles consist of hundreds to thousands.

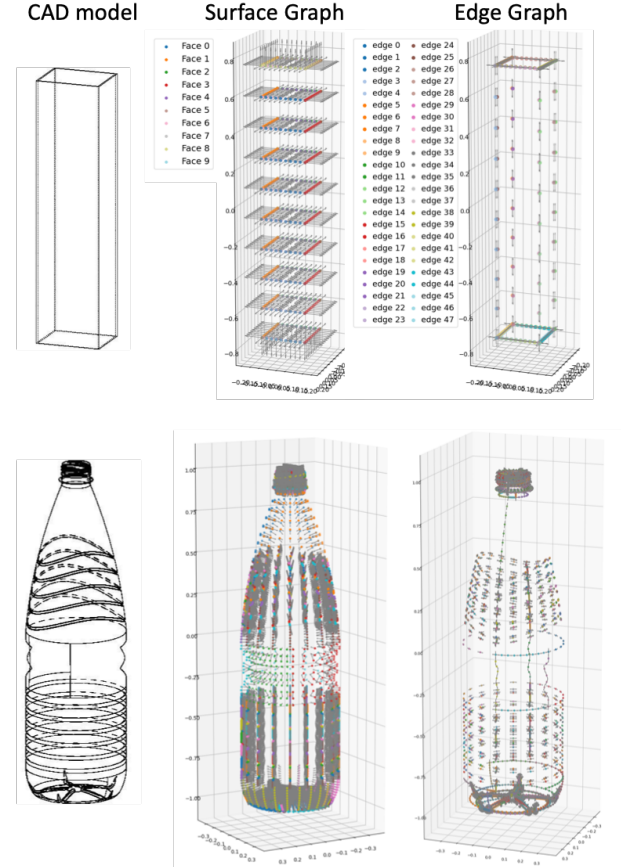

**Figure 3.** First column is the original boundary representation of the crash box and bottle, and their encoded surface graph and curves graph are shown in the second and third column.

To extract design parameter information from B-Rep without simulation data, a pseudo-task reconstructs the spatial signed distance field (SDF) via a decoder attached to the encoder's latent space. The encoder, inspired by UV-Net [7], uses a parametric graph neural network, representing curves as fixed point sets with tangent vectors and surfaces as fixed vertex grids with normal vectors. The parameterized graph representations are processed through the B-rep neural network encoder in three stages:

1. Surface Encoding: Each surface is passed through a face encoder, which utilizes a 2D CNN due to the fixed-grid representation of surfaces.
2. Edge Encoding: Similarly, edges are processed by an edge encoder, employing a 1D CNN to capture their features.
3. Graph Encoding: After inter-node message passing, the graph-level representation is constructed, where each face acts as a node and connectivity is defined by neighboring faces. The connection strength between nodes is modulated by edge features, which are outputs from the edge encoder. The graph encoder then performs message passing between nodes, followed by a pooling (readout) operation to generate a latent representation of the CAD model.

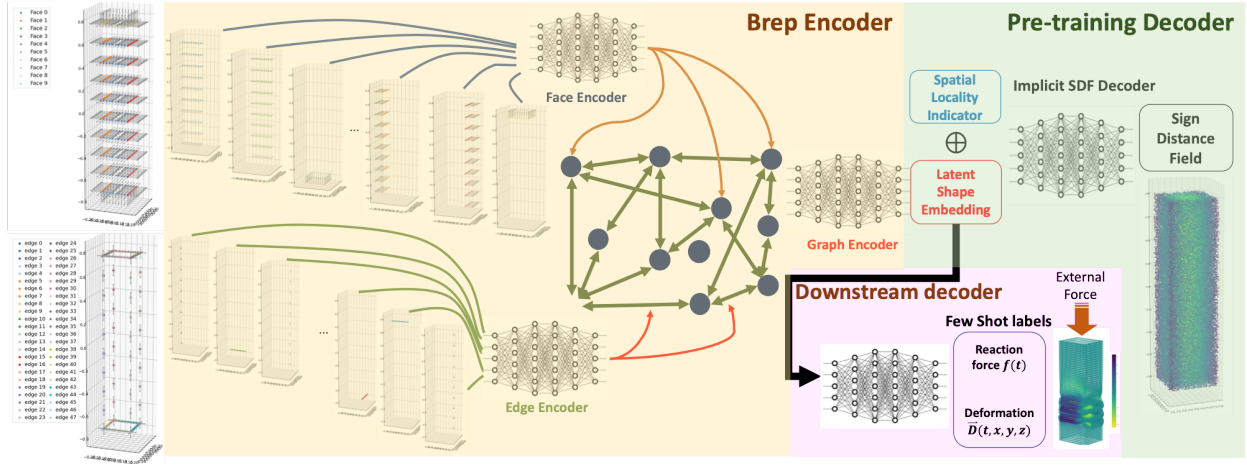

**Figure 4.** The proposed self-supervised geometric representation learning framework

During pretraining, an implicit function decoder is attached to the latent vector produced by the encoder, as illustrated in Figure 4. The decoder's task is to reconstruct the signed distance field (SDF) of the geometry. Given a shape's latent representation, it is concatenated with a point-wise spatial locality indicator and passed through fully connected layers to predict the shortest signed distance to the zero level surface at each queried location. The spatial locality features, in their simplest form, are the XYZ coordinates of the queried point—indicating where the SDF is being sampled. To provide more detailed location information to the decoder, higher-frequency components (i.e., Fourier features) of the normalized XYZ coordinates can be incorporated; their effectiveness is discussed further in Section 3.2.2.

**Table 1.** The two case studies' pretraining Hyperparameters

| Data | Encoder Size | Decoder | | | Spatial locality indicator | Sampling | Loss Function |
|---|---|---|---|---|---|---|---|
| | | Width | Depth | Size | | | |
| Crash Box | 6.07 M | 1024 | 4 | 2.31M | Normalized XYZ coordinates | Near-zero level | MSE |
| Bottle | 6.07 M | 4096 | 5 | 55.6 M | Fourier features of XYZ coordinates | Near-zero level & Batch Attention | Weighted MSE |

### 3.2.1 Near-Zero level SDF Sampling Precomputation

While precomputing a subset of the Signed Distance Field (SDF) and interpolate for coordinates that are not precomputed is essential for efficient pretraining, previous method of random sampling and uniform grid sampling strategies present major drawbacks for thin-shell geometries:
1. Excessive out-of-shape Samples – Since the volume of the shape is small relative to the bounding box, most samples are far from the surface, leading to inefficient learning [5].
2. Resolution Limitation – If the grid resolution is not at least twice as fine as the thinnest geometry feature (per Nyquist Sampling Theorem [48]), trilinear interpolation introduces inaccuracies.

To address these issues, a precomputation strategy inspired by DeepSDF [41] is adopted, biasing sample points toward the surface and its vicinity. This approach maximizes the utility of precomputed SDF values while preserving geometric details. Each crash box geometry includes 1,000,000 precomputed SDF values, while bottle geometries use 1,200,000. These quantities are sufficient to capture fine-scale geometric features as a pretraining target.

In the crash box case study, randomly selecting from the precomputed coordinate set without additional interpolation is found sufficient for the encoder to learn thickness variations effectively. Figure 5 compares two precomputed SDF fields of a crash box, each containing 64,000 points. The lower field, which prioritizes near-surface samples, results in a higher density of signed distance values near the zero level set. This improves the approximation of thin-shell geometries, ensuring the necessary accuracy for the pretraining stage to capture thickness variations.

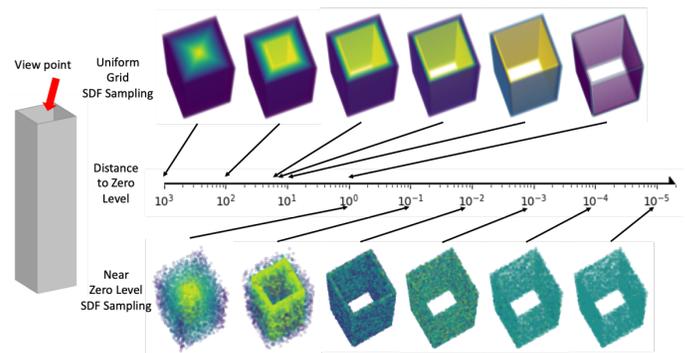

**Figure 5.** Comparison of sampling strategies and the distribution of precomputed data points within a threshold distance to the zero-level surface. The lower sampling strategy, which prioritizes near-zero level coordinates, captures a higher density of data points near the geometry's surface. This results in a more precise representation of thin-shell features, such as thickness, making it a more effective pretraining target.

The effectiveness of the proposed pretrained model is examined through two key aspects: the reconstruction quality of the decoder's output and the regression of design parameters from the encoder's output. Figure 6 presents reconstructed crash box samples from the test set. While minor imperfections exist—such as slightly rounded fillet corners and box margins that are not perfectly straight—the overall shape is well preserved.

To evaluate whether the latent space effectively encodes design parameters through pretraining, a linear probing experiment is conducted. A single linear layer, without an activation function, is applied to the latent vectors generated by the pretrained encoder to regress to the design parameters. Since this regression model relies solely on a linear projection, it provides a strict assessment of the latent space's capacity to encode design parameters. Figure 7 presents scatter plots of the test set, demonstrating that all four design parameters are successfully embedded, with test set $R^2$ scores exceeding 0.99.

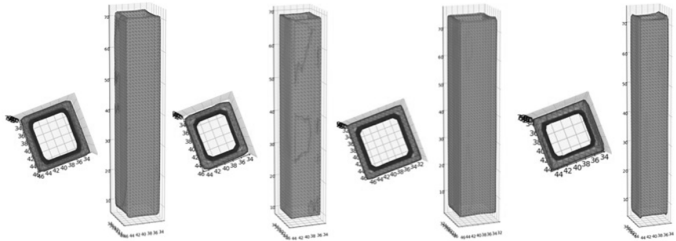

**Figure 6.** Several reconstruction results from test set. The visualized results are the marching cube results on regular grid SDF queries of the pretrained model.

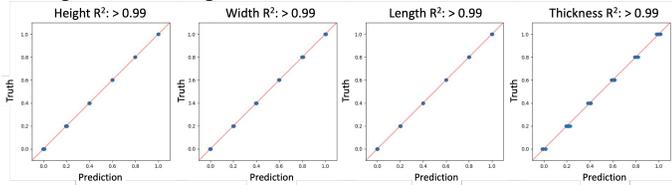

**Figure 7.** Crash boxes' linear probing regression to the four design parameters on the test set.

### 3.2.2 Periodic Locality Indicator and Batch-adaptive Attention Weighted Loss

Adapting the pretrained crash box architecture to high-fidelity bottle geometries presents challenges in capturing fine-scale features. Figure 10 shows that while large-scale curvature of the bottle's top section is preserved, intricate ribs' details remain underrepresented. To address this, two key modifications are introduced.

The first is **Fourier Feature Networks for Enhanced Locality Representation.** Coordinate-based multi-layer perceptrons (MLPs) struggle to learn high-frequency functions, a challenge known as "spectral bias" [49,50]. To enhance fine-scale reconstruction, Fourier feature network (FFN) (illustrated in Figure 8) maps spatial inputs into the frequency domain, better matching the characteristics of complex geometries, such as bottle shapes. However, this added complexity is found hindering generalization to unseen coordinates. To address this issue, a combination of random coordinate sampling (approximately 10%) and extended training iterations is used. While efficient trilinear sampling cannot be applied to the near-zero-level subset of precomputed SDFs due to their non-grid structure, this research uses a weighted average of K-nearest neighbors (KNNs) as a fast and accurate alternative for approximating the randomly selected coordinates.

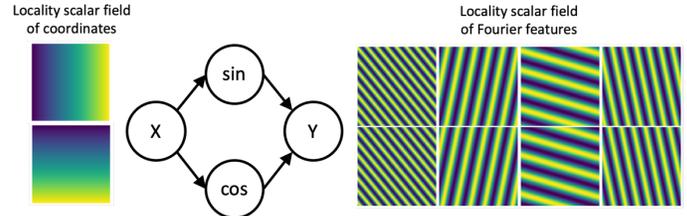

**Figure 8.** Visualization of the scalar field input and output of Fourier feature network in a 2D example.

The second is **Batch-Adaptive Attention with Loss Reweighting**. While FFN enhances static feature representation, such as the details in lower half of the bottles, dynamic features controlled by varying design parameters—namely the top section rib details in Figure 10—remain challenging. To address this, this research proposes an innovative batch-adaptive attention (BA) mechanism that adjusts the loss function to prioritize regions with significant geometric variation.

The underlying principle is straightforward: penalize SDF values that deviate significantly from others, as these variations indicate features controlled by a changing parameter. Implementation follows these steps: for each batch, precomputed SDF values from individual geometries are used to guide sampling. A subset of coordinates from one geometry's precomputed SDF set, inherently biased toward near-surface regions, is selected. The SDF of same coordinates are then queried in other geometries within the batch, approximated using weighted KNNs. This process is repeated across every geometry in the batch to create a set of shared sampled coordinates and the corresponding SDF in each geometry. Next, the batch-wise SDF difference is computed by measuring the deviation of each geometry's SDF from the batch mean at these locations. The batch-wise SDF variation is then mapped to a weight multiplier, ensuring that regions with significant geometric differences receive higher penalties—potentially up to an order of magnitude greater than stable regions. To reinforce dynamic awareness, the weight multiplier is applied to the loss function, prioritizing geometries with SDF values that deviate significantly from others at the same coordinates. This encourages the model to focus on evolving geometric features while maintaining sensitivity to globally significant structures.

The pseudo-code of this process is shown in Algorithm 1. The heatmap visualization in Figure 9 illustrates this effect: dynamic regions, where geometries exhibit noticeable differences, are highlighted, while static regions—such as the lower half of the bottle—receive lower emphasis.

**Algorithm 1.** Batch-adaptive attention weighted loss function

**Input:**
Batch size $B$
Nearest neighbor count $K$
A batch of geometries $G(b), b = 1, \ldots, B$
Number of coordinates per iteration $N_c$
Number of randomly chosen coordinates $N_r$
Number of precomputed signed distance values per geometry $N_p$
Precomputed SDF coordinates $X(b, n), n = 1, \ldots, N_p$
Signed distance values $V(b, X(b, n))$ at $X(b, n)$
Fourier feature network $F$
Pretraining model $M$

**Output:**
Loss to update the model $M$

Initialize coordinate set $C_b$ with $N_r$ random samples from $X(b, n)$ ;
**for** $b \in \{1, \ldots, B\}$ **do**
    **for** $i = 1$ to $(N_c - N_r)/B$ **do**
        Randomly select $n_b \in \{1, \ldots, N_p\}$ ;
        Retrieve coordinate $X(b, n_b)$ ;
        Retrieve true SDF value: $V(b, X(b, n_b))$ ;
        **for** $b_2 \in \{1, \ldots, B\}, b_2 \neq b$ **do**
            Approximate SDF value $V(b_2, X(b, n_b))$ using $K$-weighted nearest neighbors ;
        **end**
        Compute batch mean SDF at $X(b, n_b)$:
$$\mu(X(b, n_b)) = \frac{1}{B} \sum_{b_3=1}^{B} V(b_3, X(b, n_b))$$
        **for** $b_3 \in \{1, \ldots, B\}$ **do**
            Compute batch difference:
$$bd(b_3, X(b, n_b)) = |V(b_3, X(b, n_b)) - \mu(X(b, n_b))|$$
        **end**
        Compute mean batch difference:
$$\mu_{bd}(X(b, n_b)) = \frac{1}{B} \sum_{b_3=1}^{B} bd(b_3, X(b, n_b))$$
        **for** $b_4 \in \{1, \ldots, B\}$ **do**
            Compute batch difference weight:
$$w_{bd} = 1 + \log\left(\frac{bd(b_4, X(b, n_b)) + \mu_{bd}(X(b, n_b))}{\mu_{bd}(X(b, n_b))}\right)$$
        Predict SDF using the model:
$$\hat{V}(b_4, X(b, n_b)) = M(G(b_4), F(X(b, n_b)))$$
        Compute weighted loss:
$$L = w_{bd} \cdot (\hat{V}(b_4, X(b, n_b)) - V(b_4, X(b, n_b)))^2$$
        Update the model with the weighted loss
        **end**
    **end**
    **for** $i = 1$ to $N_r$ **do**
        Retrieve $V(b, X(b, n_b))$ using $K$-weighted nearest neighbors ;
        Apply the same weighted loss computation as above
    **end**
**end**

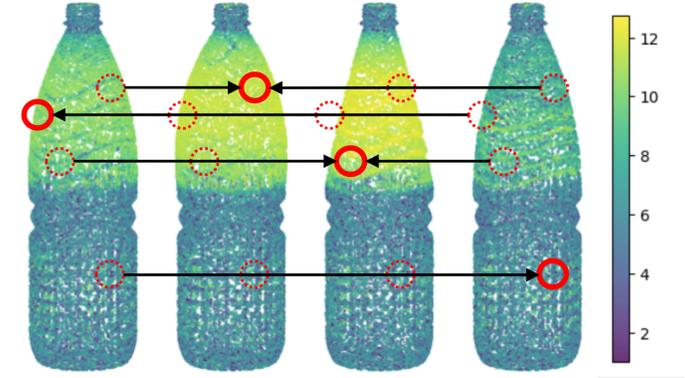

**Figure 9.** For an example batch of 4 bottles, each geometry randomly selects their existing pre-computed coordinates (solid circles) while other geometries within the same batch uses nearest neighbors to approximate and calculate the differences. The scalar field corresponds to the weight multiplier that will apply to loss function.

The effectiveness of the proposed pretrained model for bottle geometries is evaluated through reconstruction results and design parameter regression, similar to the crash box case study. Figure 10 compares the reconstruction results of the proposed method (FFN with BA) with the ground truth, DeepSDF decoder, and FFN without BA. The results show that the proposed modifications significantly improve the quality and detail of the reconstructed mesh. In contrast, DeepSDF captures only the largest-scale features of the shape, while the FFN without BA focuses on small-scale details consistent across all bottles (lower half). However, this approach fails to capture the varying design parameters controlling the upper half of the ribs.

To further validate the latent space, linear probing to regress to the normalized design parameters is again being investigated. The scatter plots and $R^2$ scores of the test set are shown in Figure 11, indicating strong parameterization, with all regression results exceeding 0.98. While the added FFN and BA components significantly improve the model's performance, thickness and pitch—parameters that affect the bottle's finer-scale geometry—still show slightly lower $R^2$ values, just below 0.99.

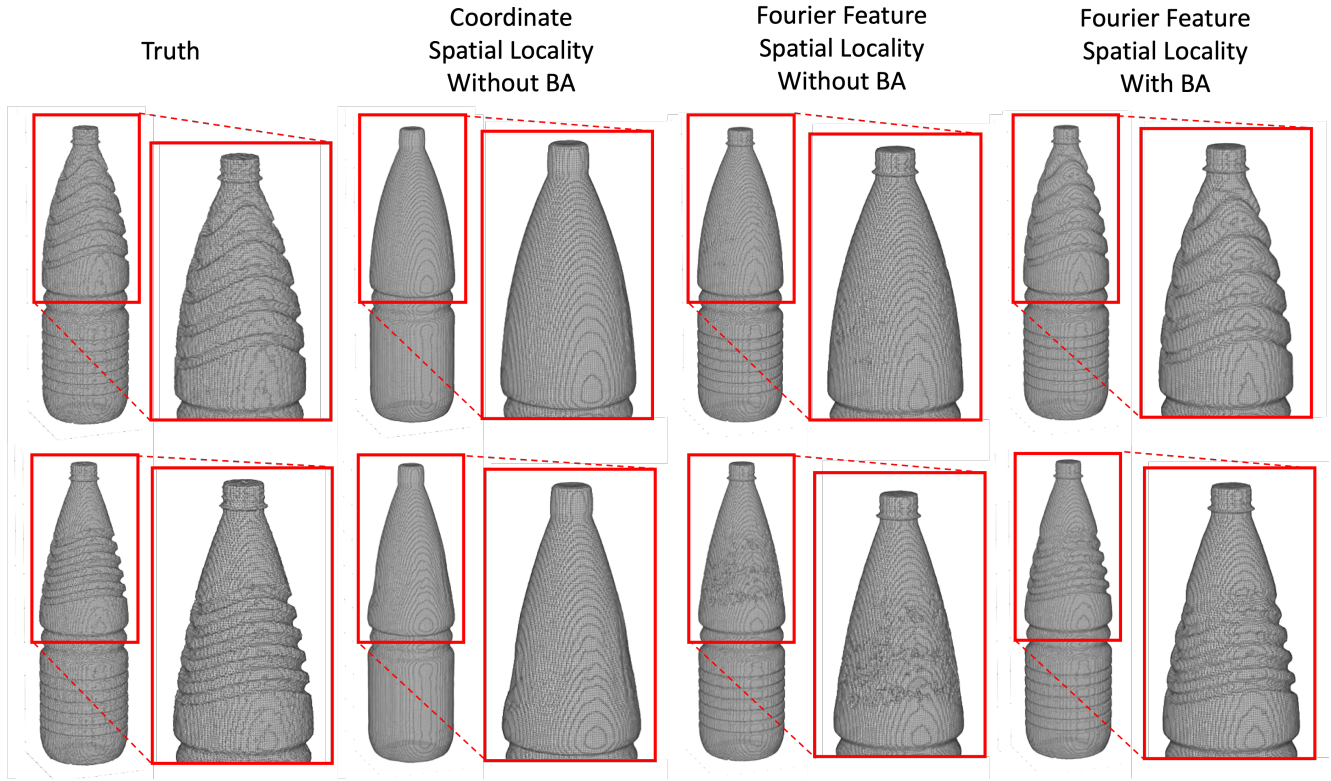

**Figure 10.** Test set results for different decoding strategies are compared to the ground truth. The visualized meshes are generated using the marching cubes algorithm based on the decoders' SDF output.

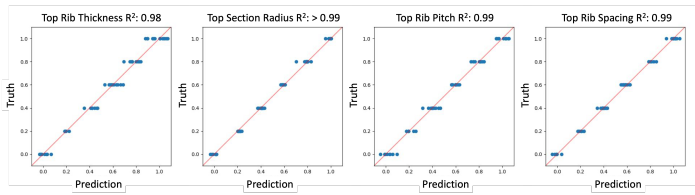

**Figure 11.** Bottles' linear probing regression to the four design parameters on the test set.

### 3.3 Simulation setup

Although pretraining aims to extract design parameters without simulation information, one still needs to generate simulations to know how the pretrained model performs compared to parametric models and training non-parametric models from scratch in downstream tasks.

For the bottle structure, the simulation is conducted using nonlinear static analysis in Abaqus/Standard, while for the crash box, dynamic analysis is performed in Abaqus/Explicit. This distinction is made because the physical events simulated differ in terms of equilibrium and dynamics. Bottle buckling is a quasi-static event with negligible mass inertia, where the total force (internal plus external) is in equilibrium. In contrast, an automotive crash is a dynamic event, characterized by non-equilibrium conditions in which the total force is equal to mass times acceleration. In the crash scenario, mass inertia, contact, and large deformations dominate the system of equations over a fraction of a second. Explicit analysis is therefore employed, using very small time increments and eliminating the need to solve the full stiffness matrix inversion.

For both case studies, the meshes are generated in Abaqus/CAE. Both the crash box and bottle structures are modeled using shell elements, each assigned a thickness and elasto-plastic material properties. For the crash box, an aluminum alloy is used, as it is a common material in automotive applications, with the following properties: density of 2700 kg/m³, elastic modulus of 70 GPa, and a Poisson's ratio of 0.33. For the bottle, a soft plastic material typically used in plastic bottles is employed. This material exhibits isotropic elasticity with a Young's modulus of 1.21 GPa, a Poisson's ratio of 0.4, and a density of 940 kg/m³. Additionally, the plastic hardening behavior is defined by a series of stress-strain data points.

Boundary conditions constrain one end of the structure by fixing all degrees of freedom, while a prescribed displacement is applied at the other end along the z-axis. Since both solvers account for geometric non-linearity, they dynamically adjusts the step size based on convergence, and therefore the effective pressing speed is not constant but instead depends on how the solver progresses through time. The reaction force fluctuates over time due to the geometry-dependent deformation response, making it a function of the structure's shape and material properties rather than a steady load application.

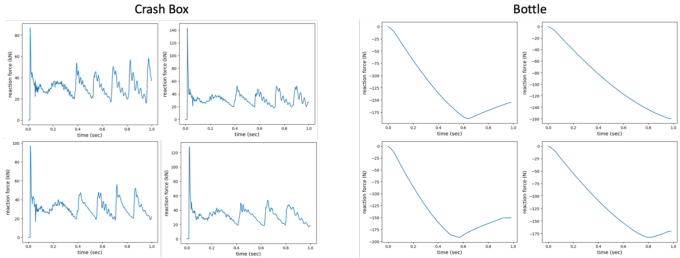

**Figure 12.** Examples of reaction force over time for four geometries from each dataset.

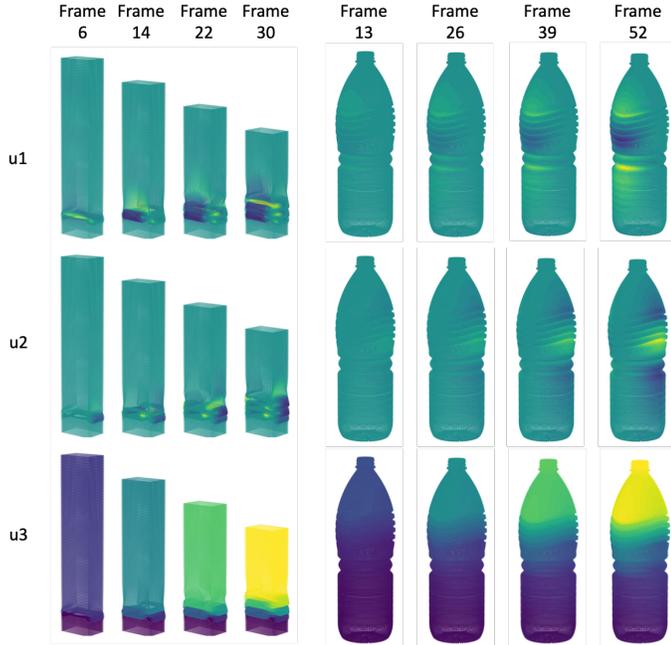

**Figure 13.** Examples of nodal deformation fields over time for one geometry from each dataset are shown. Note there are three directions of deformation for each node in each time frame.

The governing equations follow the nonlinear solution strategies that solve for equilibrium using an incremental-iterative Newton method. Various field and history outputs are recorded at each increment. This study uses two quantities to evaluate the surrogate models: reaction force over time (Figure 12), and nodal displacement field over time (Figure 13). More information can be found in the ABAQUS documentation [51].

### 3.4 Testing pretrained model on downstream tasks

The proposed geometric representation learning (GRL) method demonstrates strong parameterization of the geometries' fine-scale control design parameters. However, the primary objective of the latent vector is its performance in applications where these fine details are critical. To evaluate its effectiveness, the pretrained model is applied to downstream structural physics estimation tasks. In this process, the SDF decoder is discarded, and the encoder's latent code is integrated with a separate downstream decoder. The results are then compared to training the encoder-decoder model from scratch, which resembles non-parametric surrogate models, as well as to parametric surrogate modeling, where design parameters are directly used as inputs to the downstream decoder.

The pretrained models in two case studies are both applied to predict the reaction force over time and the temporal displacement field. To leverage the pretrained model, two approaches are proposed: (1) direct deployment of the latent representation and (2) finetuning the pretrained model.

In the direct deployment approach, the pretrained latent space is optionally normalized and then directly fed into the downstream model, mirroring how design parameters would typically be used. This is the simplest, most efficient, and ideal method for utilizing the pretrained model. For finetuning, the encoder remains learnable but is updated with a significantly lower learning rate ($0.01\times$) during downstream tasks. While this approach demands more computational resources, it provides greater flexibility and potential in adapting to task-specific outputs.

For the downstream decoder, this study uses Dassault Systèmes' proprietary model, which has previously demonstrated success when using a design parameter vector as input. To ensure a fair comparison, all models use the same downstream architecture, with the only variation being the input dimension, which aligns with the respective method. For the parametric approach, the input size is 4, corresponding to the number of varying parameters. For other methods, the input size is 64, matching the output dimension of the proposed method's GNN encoder. Most importantly, to promote few-shot learning capabilities, each setup across different training sample sizes is evaluated, ranging from tens to hundreds of data points. This analysis provides insight into the minimum data requirements necessary for effective model performance.

## 4. RESULTS
### 4.1 Temporal Reaction Force Prediction

The reaction force prediction results for the two case studies are presented in Figure 14. The mean squared error (MSE) test loss is computed for each geometry at each time frame, with varying numbers of training samples (shots). Across all cases, parametric models yield the best performance, which is expected because design parameters directly provide structured information, making them the most reliable predictors of reaction force. In low-data scenarios, using the pretrained latent representation consistently outperforms training non-parametric models from scratch, demonstrating that capturing fine-scale details during pretraining enhances few-shot structural physics performance, specifically for reaction force estimation.

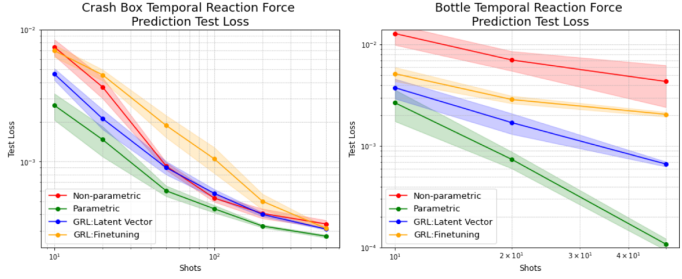

**Figure 14.** Reaction force prediction for both case studies across different numbers of shots. A "shot" refers to a complete simulation run for a given geometry, generating a full time-series of reaction force data.

However, finetuning the pretrained model does not improve reaction force prediction. In the crash box study, it even underperforms compared to both direct latent vector deployment and training non-parametric models from scratch. This is likely because the pretrained latent space is already near-optimal (Figure 7) and adjusting it with limited data distorts its learned features, leading to overfitting and a decline in predictive accuracy.

To summarize, reaction force prediction is a high-level and complex task, as it condenses the geometry's dynamic response into a single value per time frame. Findings indicate that when explicit design parameters are unavailable, directly deploying the pretrained latent representation is the most effective strategy for few-shot learning of reaction force, while finetuning degrades performance due to overfitting.

### 4.2 Temporal Deformation Field Prediction

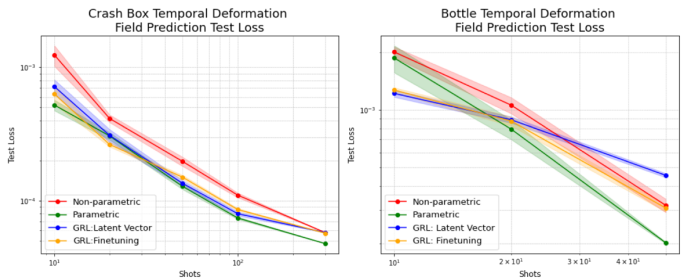

**Figure 15.** Deformation field prediction for both case studies across different numbers of shots. A "shot" refers to a complete simulation run for a given geometry, producing time-series data across multiple time frames and nodes.

The temporal nodal deformation prediction results are shown in Figure 15. The mean squared error (MSE) test loss is computed for each node, each geometry, and each time frame, across varying numbers of training samples. Due to this, the data density is significantly higher than in reaction force prediction, with 200,000 data points per geometry in the crash box case (500× more than reaction force) and 1,000,000 data points per geometry in the bottle case (20,000× more than reaction force). This difference in data availability strongly impacts model performance and the effectiveness of different learning strategies.

A key observation is that finetuning performs significantly better for displacement prediction than for reaction force prediction. This improvement is due to the nature of displacement as a more localized and continuous property, in contrast to reaction force, which represents a high-level global response. Finetuning benefits from the abundance of data available for each geometry, making it more effective in this task.

Another notable finding is that unlike in reaction force prediction—where parametric methods clearly outperform—pretrained latent representations perform comparably in both case studies, particularly in low-shot settings. This is likely because the pretraining task closely resembles the downstream task, making the latent space naturally well-aligned for deformation prediction. The pretraining process already involves mapping a latent vector to a scalar field while incorporating spatial locality indicators. In both case studies, near-surface sampling further strengthens this alignment, while in the bottle study, batch attention further enhances the model's ability to focus on dynamically relevant regions, such as the top-section ribs, where significant temporal deformation occurs (Figure 13).

While design parameters provide a compact representation of each geometry, their expressiveness may be limited due to nonlinearities in the mapping process. In contrast, the 64-dimensional latent vector, shaped by pretraining, likely captures not only the underlying design parameters but also an optimized representation that facilitates scalar field prediction. This ability to encode both design information and a more decodable mapping explains why, in the deformation case study, the latent space serves as a strong alternative to parametric methods.

Key Takeaways of this study are as follows:
- With strong reconstruction quality that captures dynamic, high-frequency details, the latent space effectively embeds fine-scale design parameters and serves as a robust alternative under few-shot scenarios.
- Higher data density and smooth, continuous targets improve finetuning performance, making it more effective than in reaction force prediction.
- When the downstream task closely aligns with the pretraining task, the latent space can even surpass parametric methods due to its expressiveness and feature richness.

### 5. CONCLUSIONS AND FUTURE WORK

Previous geometric representation learning methods have introduced training strategies that decouple non-parametric geometry encoding from corresponding labels, facilitating few-shot downstream transfer learning. However, these approaches often struggle to capture fine-scale details that are critical for structural physics prediction. This work addresses this limitation by incorporating near-zero level sampling, periodic locality indicators in input features, and a novel batch-adaptive attention mechanism into a B-Rep to SDF pretraining framework. These techniques enhance the model's ability to distinguish and reconstruct fine-scale geometric details.

The approach introduced in this paper is then validated through design parameter regression from the latent space, as well as few-shot temporal reaction force and deformation field prediction. Notably, the design parameter regression experiment demonstrates that even a simple linear probe can successfully regress to normalized design parameters, confirming that the latent space effectively captures the distribution and range of critical geometric features. Experimental results further indicate that in the few-shot regime, utilizing the pretrained latent space as a direct input consistently outperforms training non-parametric models from scratch in both accuracy and computational efficiency by eliminating the need for encoder training. Finetuning, however, may be preferable under specific conditions: when the target downstream task provides a higher density of information per geometry, when the task involves more localized and continuous properties, when the pretraining parameterization can be further optimized, or when the pretraining task structure (e.g., SDF as a scalar field) naturally aligns with the downstream objective.

For future work, while the proposed method demonstrates strong performance within controlled geometric families, its scalability to diverse CAD repositories remains an open question. Additionally, model architecture requirements vary across datasets—for example, the use of Fourier features as locality indicators results in slow convergence for crash box reconstruction with no significant performance improvement, while coordinate-based MLP leads to overly smoothed bottle reconstructions. A more automated or unified approach to architecture selection could improve robustness. Furthermore, the current encoder-decoder framework is limited to two modalities, with one requiring easy generation. Future research will focus on developing a generalized pretraining strategy that integrates multiple modalities to enhance representation quality.

## ACKNOWLEDGEMENT

This work was supported in part by the National Science Foundation under grant Award CMMI-2113301 and the Pennsylvania Infrastructure Technology Alliance. This work utilized the PSC Bridges-2 system through allocation CIS240252 from the Advanced Cyberinfrastructure Coordination Ecosystem: Services & Support (ACCESS [52]) program, supported by the National Science Foundation. grants #2138259, #2138286, #2138307, #2137603, and #2138296.

## REFERENCES

[1] Mai, H. T., Kang, J., and Lee, J., 2021, "A Machine Learning-Based Surrogate Model for Optimization of Truss Structures with Geometrically Nonlinear Behavior," Finite Elements in Analysis and Design, **196**, p. 103572. https://doi.org/10.1016/j.finel.2021.103572.

[2] Ferguson, K., Chen, Y., Chen, Y., Gillman, A., Hardin, J., and Burak Kara, L., 2024, "Topology-Agnostic Graph U-Nets for Scalar Field Prediction on Unstructured Meshes," Journal of Mechanical Design, **147**(4), p. 41701. https://doi.org/10.1115/1.4066960.

[3] Wu, H., Luo, H., Wang, H., Wang, J., and Long, M., 2024, "Transolver: A Fast Transformer Solver for PDEs on General Geometries," *Proceedings of the 41st International Conference on Machine Learning*, R. Salakhutdinov, Z. Kolter, K. Heller, A. Weller, N. Oliver, J. Scarlett, and F. Berkenkamp, eds., PMLR, pp. 53681–53705. [Online]. Available: https://proceedings.mlr.press/v235/wu24r.html.

[4] Pfaff, T., Fortunato, M., Sanchez-Gonzalez, A., and Battaglia, P., 2020, "Learning Mesh-Based Simulation with Graph Networks," *International Conference on Learning Representations*.

[5] Chen, Y., Cagan, J., and Kara, L. B., 2025, "VIRL: Volume-Informed Representation Learning towards Few-Shot Manufacturability Estimation," Journal of Intelligent Manufacturing. https://doi.org/10.1007/s10845-025-02575-8.

[6] Jones, B., Hildreth, D., Chen, D., Baran, I., Kim, V. G., and Schulz, A., 2021, "Automate: A Dataset and Learning Approach for Automatic Mating of Cad Assemblies," ACM Transactions on Graphics (TOG), **40**(6), pp. 1–18.

[7] Jayaraman, P. K., Sanghi, A., Lambourne, J. G., Willis, K. D. D., Davies, T., Shayani, H., and Morris, N., 2021, "Uv-Net: Learning from Boundary Representations," *Proceedings of the IEEE/CVF Conference on Computer Vision and Pattern Recognition*, pp. 11703–11712.

[8] Bharadwaj, A., Xu, Y., Angrish, A., Chen, Y., and Starly, B., 2019, "Development of a Pilot Manufacturing Cyberinfrastructure With an Information Rich Mechanical CAD 3D Model Repository," *MSEC2019*, Volume 1: Additive Manufacturing; Manufacturing Equipment and Systems; Bio and Sustainable Manufacturing. https://doi.org/10.1115/MSEC2019-2882.

[9] Colligan, A. R., Robinson, T. T., Nolan, D. C., Hua, Y., and Cao, W., 2022, "Hierarchical CADNet: Learning from B-Reps for Machining Feature Recognition," Computer-Aided Design, **147**, p. 103226. https://doi.org/10.1016/j.cad.2022.103226.

[10] Lambourne, J. G., Willis, K. D. D., Jayaraman, P. K., Sanghi, A., Meltzer, P., and Shayani, H., 2021, "Brepnet: A Topological Message Passing System for Solid Models," *Proceedings of the IEEE/CVF Conference on Computer Vision and Pattern Recognition*, pp. 12773–12782.

[11] Ramm, E., and Wall, W. A., 2004, "Shell Structures—a Sensitive Interrelation between Physics and Numerics," International Journal for Numerical Methods in Engineering, **60**(1), pp. 381–427. https://doi.org/10.1002/nme.967.

[12] Kim, N. H., and Chang, Y., 2005, "Eulerian Shape Design Sensitivity Analysis and Optimization with a Fixed Grid," Computer Methods in Applied Mechanics and

Engineering, **194**(30), pp. 3291–3314. https://doi.org/10.1016/j.cma.2004.12.019.

[13] Shahrjerdi, A., and Bahramibabamiri, B., 2015, "The Effect of Different Geometrical Imperfection of Buckling of Composite Cylindrical Shells Subjected to Axial Loading," International Journal of Mechanical and Materials Engineering, **10**(1), p. 6. https://doi.org/10.1186/s40712-015-0033-z.

[14] Jones, B. T., Hu, M., Kodnongbua, M., Kim, V. G., and Schulz, A., 2023, *Self-Supervised Representation Learning for CAD*.

[15] Wong, J. C., Ooi, C. C., Chattoraj, J., Lestandi, L., Dong, G., Kizhakkinan, U., Rosen, D. W., Jhon, M. H., and Dao, M. H., 2022, "Graph Neural Network Based Surrogate Model of Physics Simulations for Geometry Design," *2022 IEEE Symposium Series on Computational Intelligence (SSCI)*, pp. 1469–1475. https://doi.org/10.1109/SSCI51031.2022.10022022.

[16] Alizadeh, R., Allen, J. K., and Mistree, F., 2020, "Managing Computational Complexity Using Surrogate Models: A Critical Review," Research in Engineering Design, **31**(3), pp. 275–298. https://doi.org/10.1007/s00163-020-00336-7.

[17] Mozaffar, M., Bostanabad, R., Chen, W., Ehmann, K., Cao, J., and Bessa, M. A., 2019, "Deep Learning Predicts Path-Dependent Plasticity," Proceedings of the National Academy of Sciences, **116**(52), pp. 26414–26420. https://doi.org/10.1073/pnas.1911815116.

[18] Truong, T. T., Dinh-Cong, D., Lee, J., and Nguyen-Thoi, T., 2020, "An Effective Deep Feedforward Neural Networks (DFNN) Method for Damage Identification of Truss Structures Using Noisy Incomplete Modal Data," Journal of Building Engineering, **30**, p. 101244. https://doi.org/10.1016/j.jobe.2020.101244.

[19] Yun, C.-B., and Bahng, E. Y., 2000, "Substructural Identification Using Neural Networks," Computers & Structures, **77**(1), pp. 41–52. https://doi.org/10.1016/S0045-7949(99)00199-6.

[20] Ribeiro, J. P. A., Tavares, S. M. O., and Parente, M., 2021, "Stress–Strain Evaluation of Structural Parts Using Artificial Neural Networks," Proceedings of the Institution of Mechanical Engineers, Part L, **235**(6), pp. 1271–1286. https://doi.org/10.1177/1464420721992445.

[21] Kazeruni, M., and Ince, A., 2023, "Data-Driven Artificial Neural Network for Elastic Plastic Stress and Strain Computation for Notched Bodies," Theoretical and Applied Fracture Mechanics, **125**, p. 103917. https://doi.org/10.1016/j.tafmec.2023.103917.

[22] Kudela, J., and Matousek, R., 2022, "Recent Advances and Applications of Surrogate Models for Finite Element Method Computations: A Review," Soft Computing, **26**(24), pp. 13709–13733. https://doi.org/10.1007/s00500-022-07362-8.

[23] Nie, Z., Jiang, H., and Kara, L. B., 2020, "Stress Field Prediction in Cantilevered Structures Using Convolutional Neural Networks," Journal of Computing and Information Science in Engineering, **20**(1). https://doi.org/10.1115/1.4044097.

[24] Jiang, H., Nie, Z., Yeo, R., Farimani, A. B., and Kara, L. B., 2021, "StressGAN: A Generative Deep Learning Model for Two-Dimensional Stress Distribution Prediction," Journal of Applied Mechanics, Transactions ASME, **88**(5). https://doi.org/10.1115/1.4049805.

[25] Ferguson, K., Gillman, A., Hardin, J., and Kara, L. B., 2024, "Scalar Field Prediction on Meshes Using Interpolated Multiresolution Convolutional Neural Networks," Journal of Applied Mechanics, **91**(10), p. 101002. https://doi.org/10.1115/1.4065782.

[26] Ning, F., Shi, Y., Cai, M., Xu, W., and Zhang, X., 2020, "Manufacturing Cost Estimation Based on a Deep-Learning Method," Journal of Manufacturing Systems, **54**, pp. 186–195. https://doi.org/10.1016/j.jmsy.2019.12.005.

[27] Yoo, S., and Kang, N., 2021, "Explainable Artificial Intelligence for Manufacturing Cost Estimation and Machining Feature Visualization," Expert Systems with Applications, **183**, p. 115430. https://doi.org/10.1016/j.eswa.2021.115430.

[28] Williams, G., Meisel, N. A., Simpson, T. W., and McComb, C., 2019, "Design Repository Effectiveness for 3D Convolutional Neural Networks: Application to Additive Manufacturing," Journal of Mechanical Design, **141**(11), p. 111701. https://doi.org/10.1115/1.4044199.

[29] Dering, M. L., and Tucker, C. S., 2017, "A Convolutional Neural Network Model for Predicting a Product's Function, Given Its Form," Journal of Mechanical Design, **139**(11), p. 111408. https://doi.org/10.1115/1.4037309.

[30] Pathak, D., Krähenbühl, P., Donahue, J., Darrell, T., and Efros, A. A., 2016, "Context Encoders: Feature Learning by Inpainting," *Proceedings of the IEEE Conference on Computer Vision and Pattern Recognition*, pp. 2536–2544.

[31] He, K., Chen, X., Xie, S., Li, Y., Dollár, P., and Girshick, R., 2022, "Masked Autoencoders Are Scalable Vision Learners," *Proceedings of the IEEE/CVF Conference on Computer Vision and Pattern Recognition (CVPR)*, pp. 16000–16009.

[32] Gidaris, S., Singh, P., and Komodakis, N., 2018, "Unsupervised Representation Learning by Predicting Image Rotations," *International Conference on Learning Representations*. [Online]. Available: http://arxiv.org/abs/1803.07728.

[33] Chen, T., Kornblith, S., Norouzi, M., and Hinton, G., 2020, "A Simple Framework for Contrastive Learning of Visual Representations," *Proceedings of the 37th International Conference on Machine Learning*, H.D. III, and A. Singh, eds., PMLR, pp. 1597–1607. [Online]. Available: https://proceedings.mlr.press/v119/chen20j.html.

[34] Brown, T. B., Mann, B., Ryder, N., Subbiah, M., Kaplan, J., Dhariwal, P., Neelakantan, A., Shyam, P., Sastry, G., Askell, A., Agarwal, S., Herbert-Voss, A., Krueger, G., Henighan, T., Child, R., Ramesh, A., Ziegler, D. M., Wu, J., Winter, C., Hesse, C., Chen, M., Sigler, E., Litwin, M.,

Gray, S., Chess, B., Clark, J., Berner, C., Mccandlish, S., Radford, A., Sutskever, I., and Amodei, D., 2020, "Language Models Are Few-Shot Learners," Advances in neural information processing systems, **33**, pp. 1877–1901.

[35] Liu, Y., Ott, M., Goyal, N., Du, J., Joshi, M., Chen, D., Levy, O., Lewis, M., Zettlemoyer, L., and Stoyanov, V., 2019, "Roberta: A Robustly Optimized Bert Pretraining Approach," arXiv preprint arXiv:1907.11692.

[36] Openai, A. R., Openai, K. N., Openai, T. S., and Openai, I. S., 2018, *Improving Language Understanding by Generative Pre-Training*. [Online]. Available: https://gluebenchmark.com/leaderboard.

[37] Radford, A., Wu, J., Child, R., Luan, D., Amodei, D., Sutskever, I., and others, 2019, "Language Models Are Unsupervised Multitask Learners," OpenAI blog, **1**(8), p. 9.

[38] Xie, L., Lu, Y., Furuhata, T., Yamakawa, S., Zhang, W., Regmi, A., Kara, L., and Shimada, K., 2022, "Graph Neural Network-Enabled Manufacturing Method Classification from Engineering Drawings," Computers in Industry, **142**. https://doi.org/10.1016/j.compind.2022.103697.

[39] Zhang, W., Joseph, J., Yin, Y., Xie, L., Furuhata, T., Yamakawa, S., Shimada, K., and Kara, L. B., 2023, "Component Segmentation of Engineering Drawings Using Graph Convolutional Networks," Computers in Industry, **147**, p. 103885. https://doi.org/10.1016/j.compind.2023.103885.

[40] Chen, Y., Kara, L. B., and Cagan, J., 2023, "BIGNet: A Deep Learning Architecture for Brand Recognition with Geometry-Based Explainability," Journal of Mechanical Design, **146**(5), p. 51701. https://doi.org/10.1115/1.4063760.

[41] Park, J. J., Florence, P., Straub, J., Newcombe, R., and Lovegrove, S., 2019, "DeepSDF: Learning Continuous Signed Distance Functions for Shape Representation," *Proceedings of the IEEE/CVF Conference on Computer Vision and Pattern Recognition*, pp. 165–174.

[42] Sitzmann, V., Martel, J., Bergman, A., Lindell, D., and Wetzstein, G., 2020, "Implicit Neural Representations with Periodic Activation Functions," Advances in neural information processing systems, **33**, pp. 7462–7473.

[43] Tancik, M., Srinivasan, P., Mildenhall, B., Fridovich-Keil, S., Raghavan, N., Singhal, U., Ramamoorthi, R., Barron, J., and Ng, R., 2020, "Fourier Features Let Networks Learn High Frequency Functions in Low Dimensional Domains," Advances in neural information processing systems, **33**, pp. 7537–7547.

[44] Lindell, D. B., Van Veen, D., Park, J. J., and Wetzstein, G., 2022, "Bacon: Band-Limited Coordinate Networks for Multiscale Scene Representation," *Proceedings of the IEEE/CVF Conference on Computer Vision and Pattern Recognition*, pp. 16252–16262.

[45] Bi, J., Ngo Ngoc, C., Yao, J., and Oancea, V., 2023, "Towards 3d Interactive Design Exploration via Neural Networks," *NAFEMS World Congress 2023*.

[46] Abdullah, N. A. Z., Sani, M. S. M., Salwani, M. S., and Husain, N. A., 2020, "A Review on Crashworthiness Studies of Crash Box Structure," Thin-Walled Structures, **153**, p. 106795. https://doi.org/10.1016/j.tws.2020.106795.

[47] Keawjaroen, P., and Suvanjumrat, C., 2017, "A Master Shape of Bottles for Design under Desirable Geometry and Top Load Test," *MATEC Web of Conferences*, p. 2008.

[48] Shannon, C. E., 1949, "Communication in the Presence of Noise," Proceedings of the IRE, **37**(1), pp. 10–21. https://doi.org/10.1109/JRPROC.1949.232969.

[49] Basri, R., Galun, M., Geifman, A., Jacobs, D., Kasten, Y., and Kritchman, S., 2020, "Frequency Bias in Neural Networks for Input of Non-Uniform Density," *Proceedings of the 37th International Conference on Machine Learning*, H.D. III, and A. Singh, eds., PMLR, pp. 685–694. [Online]. Available: https://proceedings.mlr.press/v119/basri20a.html.

[50] Rahaman, N., Baratin, A., Arpit, D., Draxler, F., Lin, M., Hamprecht, F., Bengio, Y., and Courville, A., 2019, "On the Spectral Bias of Neural Networks," *Proceedings of the 36th International Conference on Machine Learning*, K. Chaudhuri, and R. Salakhutdinov, eds., PMLR, pp. 5301–5310. [Online]. Available: https://proceedings.mlr.press/v97/rahaman19a.html.

[51] 2025, *Abaqus Analysis User's Manual*, Dassault Systèmes Corp., Providence, RI, USA. [Online]. Available: https://www.3ds.com/support/documentation/user-guides.

[52] Boerner, T. J., Deems, S., Furlani, T. R., Knuth, S. L., and Towns, J., 2023, "Access: Advancing Innovation: Nsf's Advanced Cyberinfrastructure Coordination Ecosystem: Services & Support," *Practice and Experience in Advanced Research Computing 2023: Computing for the Common Good*, pp. 173–176.

# Supplemental Material: Attention to Detail: Fine-Scale Feature Preservation-Oriented Geometric Pre-training for AI-Driven Surrogate Modeling

## A. Benchmark design parameter regression with other geometric pretraining methods

To quantitatively demonstrate the effectiveness of the proposed geometric pretraining method, it is benchmarked against previous approaches by evaluating how well the latent vectors are able to regress to the design parameters with linear probing, following the same evaluation process shown in Figure 7 and 11. For a fair comparison, all encoders in this section are scaled to have a similar number of learnable parameters, approximately 6.1 million.

Table 2 compares the proposed method with UVNet-SSL [7], surface rendering [14], and VIRL [5]. The results indicate that the proposed method outperforms all existing approaches across all four design parameters in terms of both $R^2$ score.

**Table 2.** $R^2$ scores for the four design parameters of the crash box across different self-supervised pretraining methods

| Pretrain methods | Height | Width | Length | Thickness |
|---|---|---|---|---|
| Proposed GRL | > 0.99 | > 0.99 | > 0.99 | > 0.99 |
| VIRL | 0.74 | 0.80 | 0.98 | 0.31 |
| Surface Rendering | > 0.99 | > 0.99 | > 0.99 | 0.23 |
| UVNet-SSL | 0.91 | 0.96 | 0.97 | < 0.01 |

For the bottle case study, the increased geometric complexity and the limitations of VIRL and surface rendering prevent their direct application. These methods require a fixed-length encoding for the B-Rep, restricting faces and curves to predefined types such as planes, cylinders, cones, and tori for surfaces, and lines, circles, and ellipses for curves. Since the geometric components of the bottles cannot be represented within this encoding scheme, these methods are not applicable. Instead, benchmarking is conducted using coordinate-based spatial locality information as input, as well as Fourier features without batch-adaptive attention weight adjustment, as shown in Table 3. The results demonstrate the incremental benefits of these two components. Specifically, replacing coordinate inputs with Fourier features leads to a significant performance improvement for thickness, pitch, and spacing. Furthermore, incorporating batch-adaptive attention into the Fourier feature network enables the $R^2$ score for all four design parameters to exceed 0.98.

**Table 3.** $R^2$ scores for the four design parameters of the bottle across different self-supervised pretraining methods. Note that the coordinate spatial locality method corresponds to the approach used in the crash box case study.

| Pretrain methods | Thickness | Radius | Pitch | Spacing |
|---|---|---|---|---|
| Proposed GRL | 0.98 | > 0.99 | 0.99 | > 0.99 |
| Fourier Feature Spatial Locality | 0.96 | > 0.99 | 0.22 | 0.91 |
| Coordinate Spatial Locality (crash box) | 0.96 | > 0.99 | 0.17 | 0.86 |
| UVNet-SSL | < 0.01 | 0.70 | < 0.01 | < 0.01 |

## B. Downstream task visualization

This section presents visualization results for the downstream tasks. Randomly selected test set samples are shown for both reaction force and displacement field predictions. The number of shots visualized in each case study is selected to highlight performance differences among the compared methods. Providing too much data results in similar performance across methods, while too little data prevents any method from working effectively. The corresponding data quantities are detailed in Table 4.

For reaction force prediction, Figures 16 and 17 illustrate the results for the crash box and bottle, respectively. For displacement field visualization, rather than displaying the field directly, the normalized difference between the ground truth and the predicted results is shown instead. Figures 18 and 19 present these error visualizations for the crash box and bottle, respectively. For all case studies, the pretrained latent vector demonstrates performance most comparable to parametric methods, where design parameters are directly used as input.

**Table 4:** Overview of the simulation data used for visualization.

| geometry | task | frames | nodes | shots | total data |
|---|---|---|---|---|---|
| crash box | reaction force | 450 | NA | 20 | 9000 |
| bottle | reaction force | 55 | NA | 50 | 2750 |
| crash box | displacement field | 30 | 8000 | 50 | 12000000 |
| bottle | displacement field | 55 | 17000 | 10 | 9350000 |

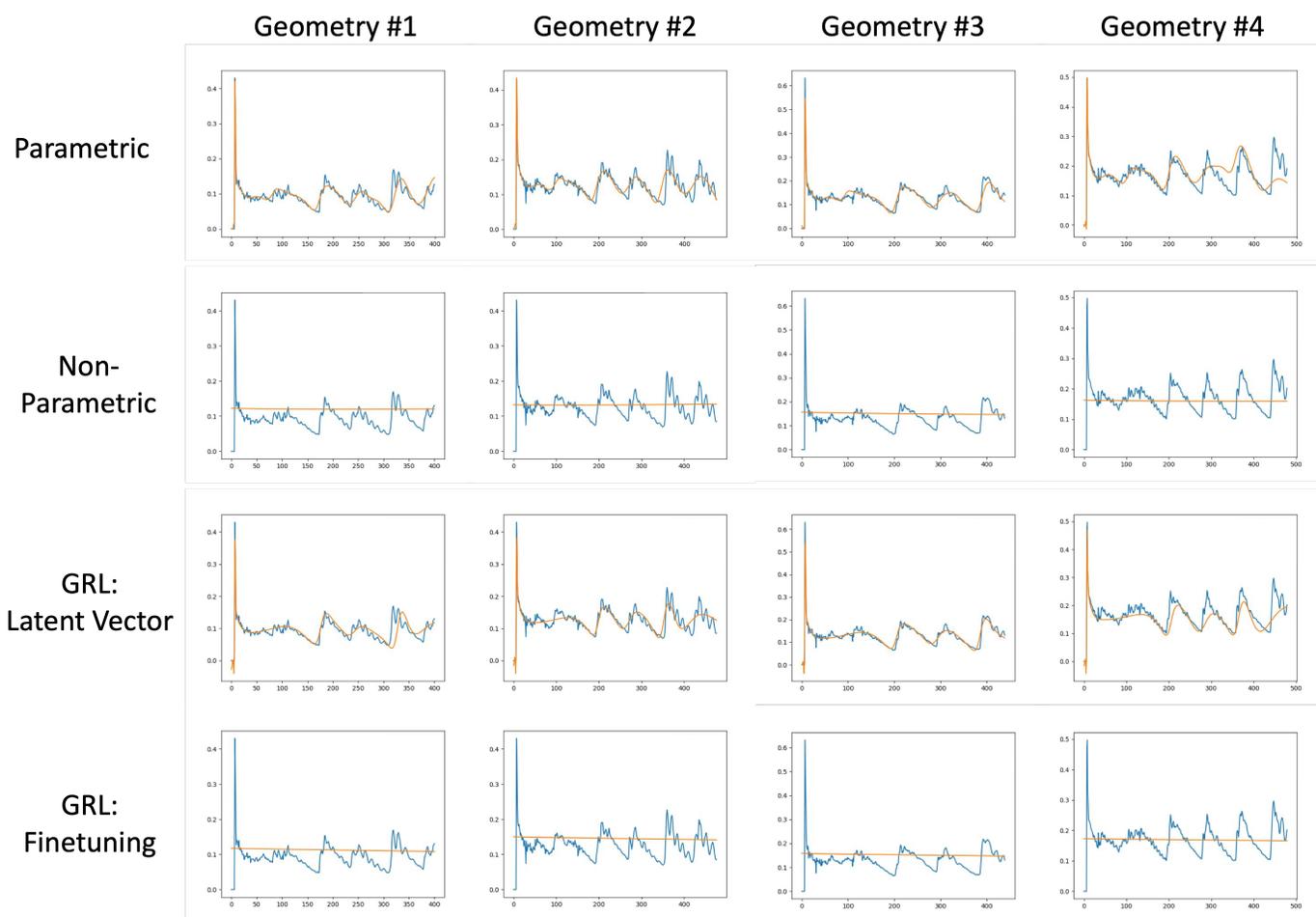

**Figure 16.** Crash box reaction force prediction of randomly selected test samples. The blue curve represents the ground truth, while the orange curve represents the prediction.

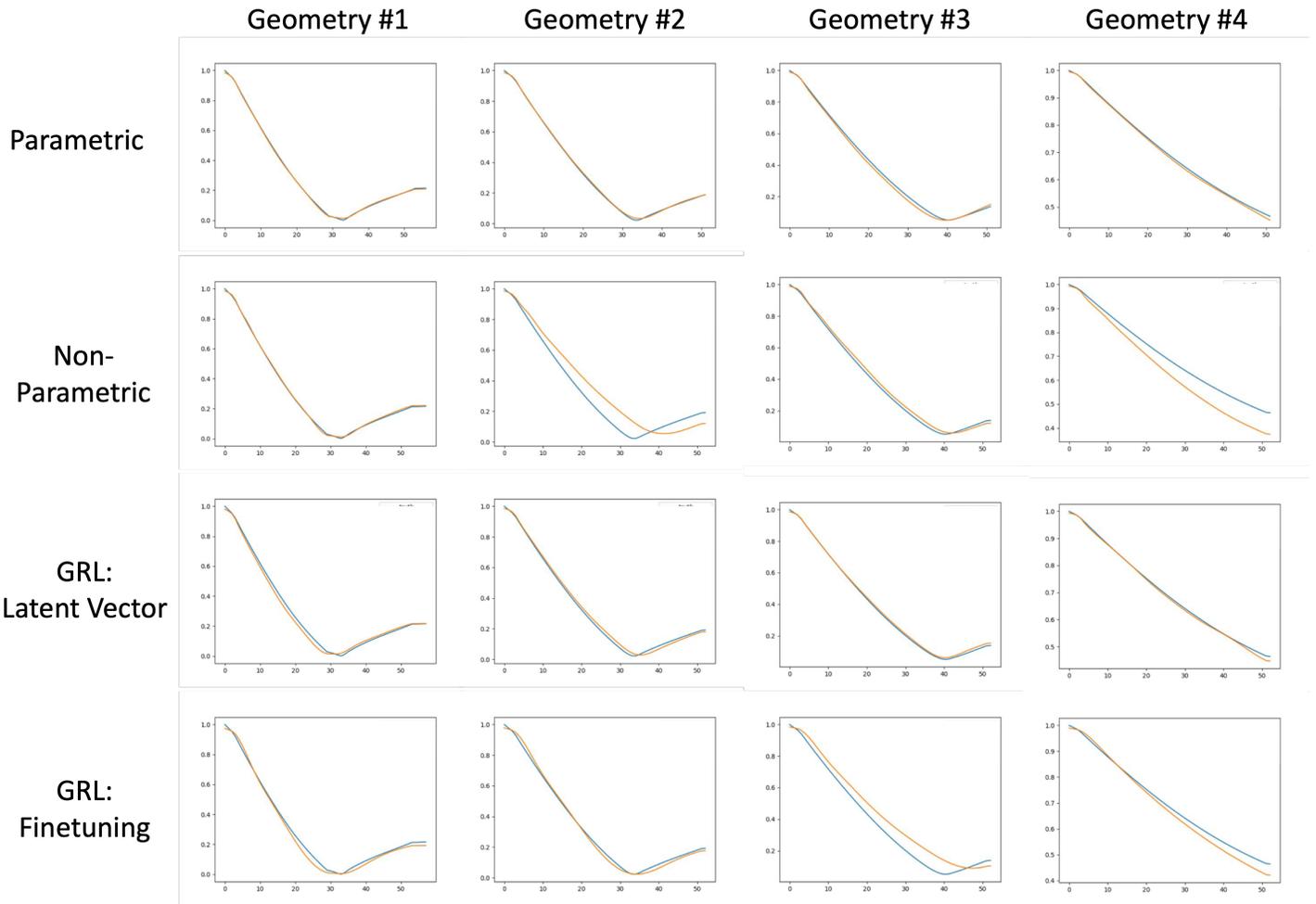

**Figure 17.** Bottle reaction force prediction of selected test samples. The blue curve represents the ground truth, while the orange curve represents the prediction.

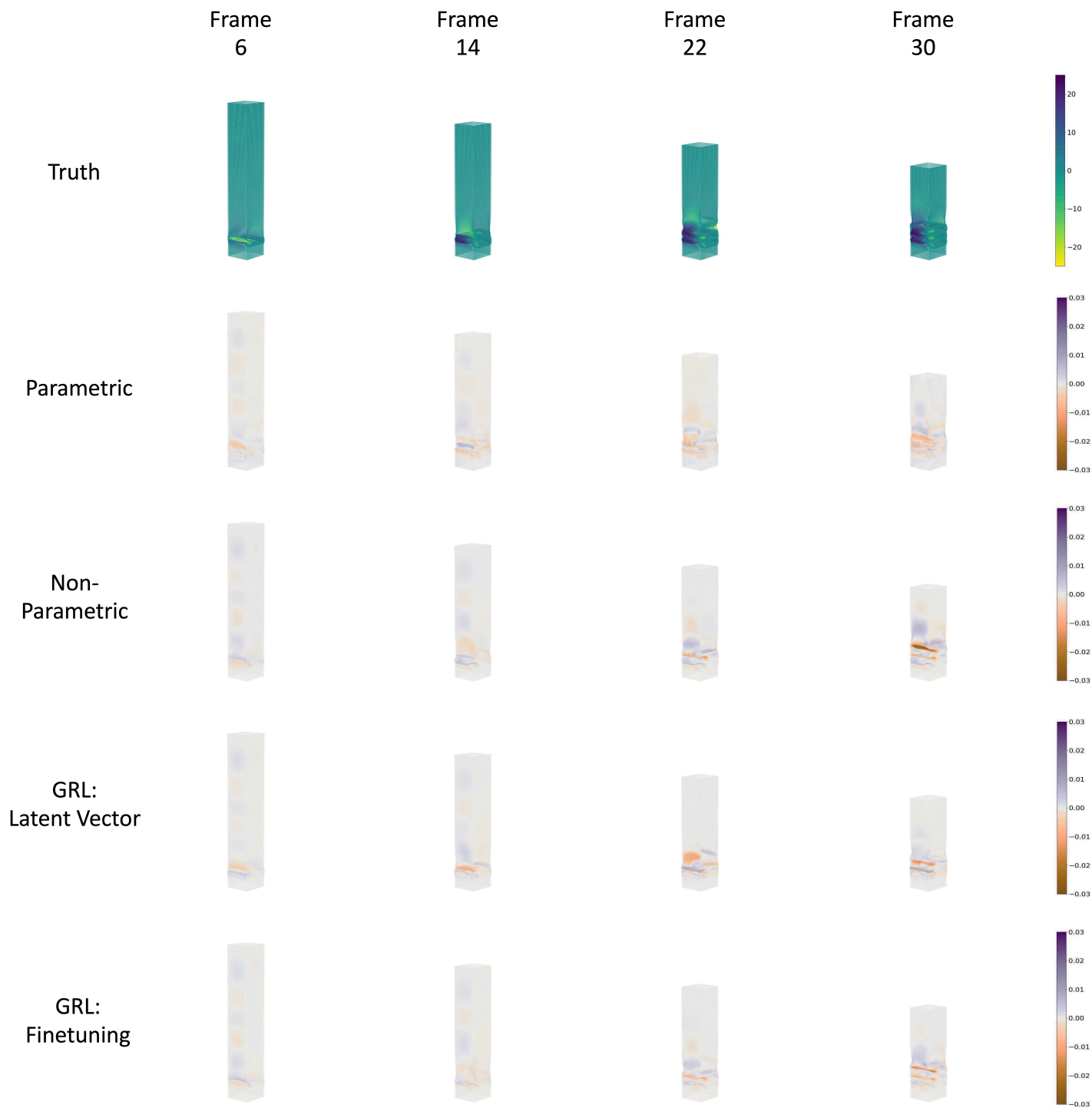

**Figure 18.** Error field visualization for crash box displacement field prediction on a test sample. The values represent the difference between the ground truth and the prediction, normalized by the nominal value of the geometry. Purple indicates positive error, while orange indicates negative error.

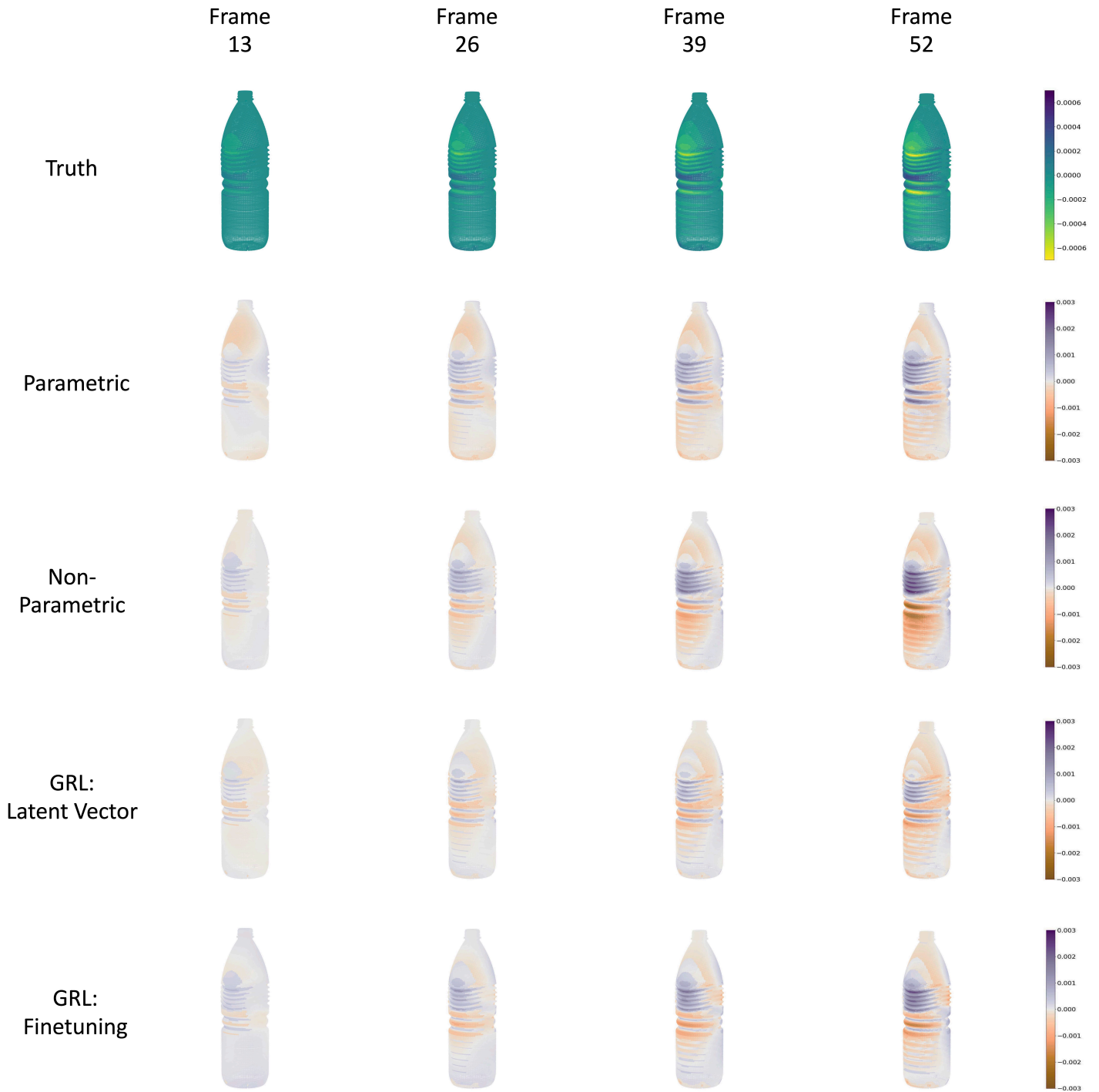

**Figure 19.** Error field visualization for bottle displacement field prediction on a test sample. The values represent the difference between the ground truth and the prediction, normalized by the nominal value of the geometry. Purple indicates positive error, while orange indicates negative error.